\documentclass[manuscript,authorversion,screen,nonacm]{acmart}

\AtBeginDocument{%
  \providecommand\BibTeX{{%
    \normalfont B\kern-0.5em{\scshape i\kern-0.25em b}\kern-0.8em\TeX}}}

\setcopyright{acmcopyright}
\copyrightyear{2018}
\acmYear{2018}
\acmDOI{XXXXXXX.XXXXXXX}

\acmConference[ETRA '24]{ACM Symposium on Eye Tracking Research \% Applications}{June 04--07,
  2024}{Glasgow, United Kingdom}
\acmPrice{15.00}
\acmISBN{978-1-4503-XXXX-X/18/06}

\usepackage{graphicx}
\usepackage{amsmath}
\usepackage{booktabs}
\usepackage{tabularx}
\usepackage{anyfontsize}
\usepackage{color, colortbl}
\usepackage{subcaption}

\usepackage[capitalize]{cleveref}
\crefname{section}{Sec.}{Secs.}
\Crefname{section}{Section}{Sections}
\Crefname{table}{Table}{Tables}
\crefname{table}{Tab.}{Tabs.}
\Crefname{subsection}{Subsection}{Subsections}
\crefname{subsection}{Subsec.}{Subsecs.}
\Crefname{equation}{Equation}{Equations}
\crefname{equation}{Eq.}{Eqs.}

\usepackage{todonotes}

\begin{teaserfigure}
    \centering
    \includegraphics[width=\columnwidth]{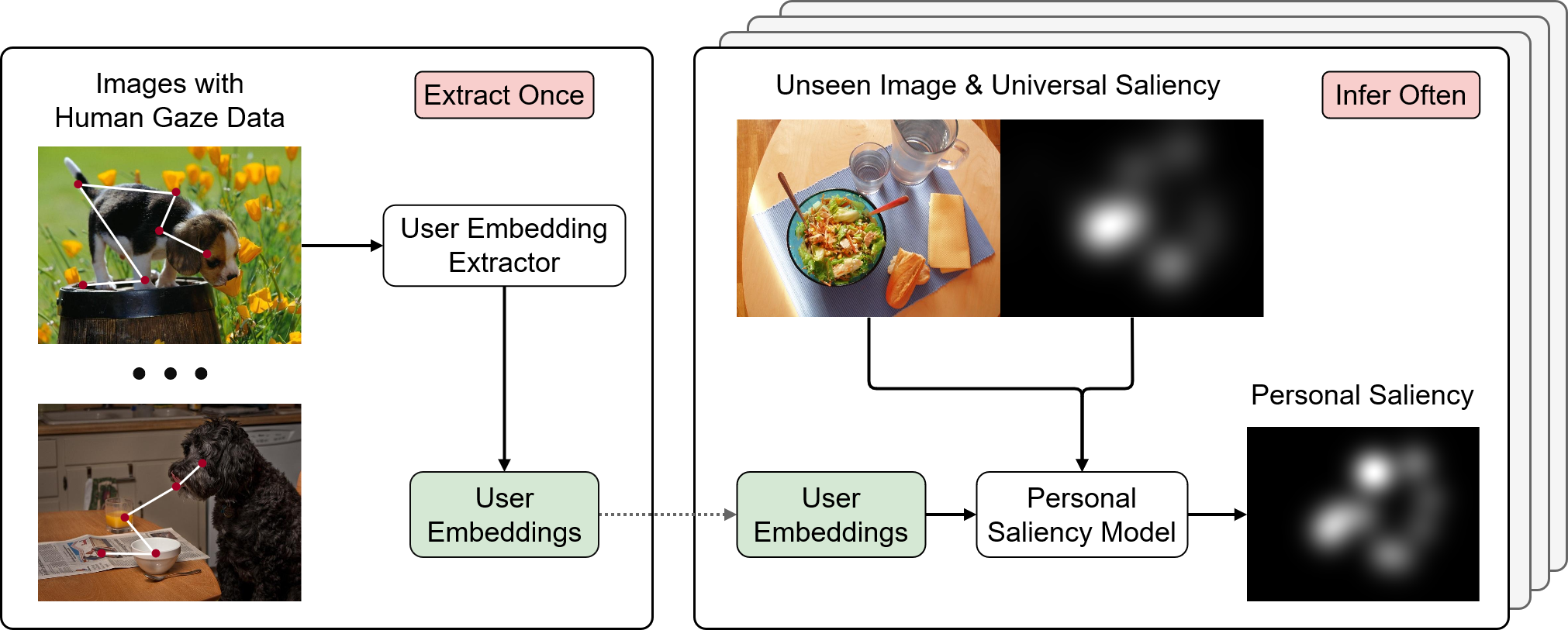}
    \caption{
        Our method (left) takes as input a small set of images and their corresponding user-specific saliency maps obtained from human gaze recorded using a stationary eye tracker and produces one user embedding for each user, which captures the user-specific differences in viewing behaviour. We then demonstrate (right) how the learned embeddings can be used to refine a universal saliency map (e.g. obtained from a saliency predictor) to an individual, personal saliency map for any new images.
    }
    \label{fig:teaser}
\end{teaserfigure}

\begin{document}

\title{Learning User Embeddings from Human Gaze for Personalised Saliency Prediction}

\author{Florian Strohm}
\email{florian.strohm@vis.uni-stuttgart.de}
\orcid{0000-0002-3787-3062}
\affiliation{%
  \institution{University of Stuttgart}
  \country{Germany}
}

\author{Mihai Bâce}
\orcid{0000-0002-1446-379X}
\affiliation{%
  \institution{KU Leuven}
  \country{Belgium}
}
\authornote{work conducted while at University of Stuttgart}
\email{mihai.bace@kuleuven.be}

\author{Andreas Bulling}
\orcid{0000-0001-6317-7303}
\affiliation{%
  \institution{University of Stuttgart}
  \country{Germany}
}
\email{andreas.bulling@vis.uni-stuttgart.de}

\renewcommand{\shortauthors}{Strohm et al.}

\begin{abstract}

Reusable embeddings of user behaviour have shown significant performance improvements for the personalised saliency prediction task. 
However, prior works require explicit user characteristics and preferences as input, which are often difficult to obtain.  
We present a novel method to extract user embeddings from pairs of natural images and corresponding saliency maps generated from a small amount of user-specific eye tracking data.
At the core of our method is a Siamese convolutional neural encoder that learns the user embeddings by contrasting the image and personal saliency map pairs of different users. 
Evaluations on two public saliency datasets show that the generated embeddings have high discriminative power, are effective at refining universal saliency maps to the individual users, and generalise well across users and images. 
Finally, based on our model's ability to encode individual user characteristics, our work points towards other applications that can benefit from reusable embeddings of gaze behaviour.

\end{abstract}

\ccsdesc[500]{Human-centered computing~User models}
\ccsdesc[300]{Computing methodologies~Interest point and salient region detections}

\keywords{gaze, eye-tracking, saliency, personal saliency, user embeddings, user model, deep learning}

\maketitle

\section{Introduction}
Saliency prediction is the task of identifying salient regions within an image which are likely to attract gaze. 
Various models have been developed which take into account both low-level features~\cite{itti2001computational, itti1998model, walther2006modeling} and high-level image characteristics~\cite{linardos2021deepgaze, pan2016shallow, fosco2020much}, incorporating bottom-up attention mechanisms, as well as task demands~\cite{mondal2023gazeformer, chen2021predicting, yang2020predicting}, which involve top-down attention processes.
Given the potential for anticipating user attention, saliency prediction models have had significant impact in computer vision and beyond, and have proven highly beneficial for a wide range of tasks, from serving as an inductive bias for neural attention mechanisms~\cite{sood2021vqa, he2019exploring, sood23_gaze} to estimating users' cognitive states~\cite{wachowiak2022analysing, iqbal2004using, abdou2022gaze}, or enabling personalised predictions for various human-computer-interaction tasks~\cite{vertegaal2002designing,zhao2016gaze,ma2002user,ardizzone2013saliency,he2019human}.

A large body of work on saliency modelling has focused on \textit{universal saliency}, i.e. the task of predicting saliency maps that aggregate gaze data from multiple observers and, as such, disregard individual differences in viewing behaviour.
There are, however, significant individual variations in how visual attention is deployed on image stimuli that are
due to a range of factors, such as scene complexity and semantics~\cite{yarbus1967eye, de2019individual}, level of expertise~\cite{buswell1935people, silva2022differences, eivazi2012gaze, brams2019relationship}, age~\cite{yu2018personalization}, or personality traits~\cite{baranes2015eye, risko2012curious}.

Despite these differences among individuals and the many applications (e.g. assistive systems~\cite{strohm2021neural, strohm2023facial, sattar2020deep}) that could benefit from a better understanding of the individual, only few previous works have proposed methods to predict \textit{personalised saliency}.
One approach involved
training an individual (sub-) model for each user, which lacks generalisability~\cite{xu2018personalized, moroto2020few, xu2017beyond}. 
Another one leverages person-specific information such as age, gender, or preference towards specific object categories or colours~\cite{xu2018personalized}. 
However, in addition to raising privacy concerns, collecting these characteristics is tedious and requires explicit user input. 

In contrast to explicitly collecting user information, we introduce a novel method to extract embeddings from users' gaze behaviour while viewing natural images. 
Our method uses a Siamese convolutional neural encoder that takes multiple images and their corresponding saliency maps of a particular user as input and produces a user embedding as output.
The embedding is learned by contrasting input pairs from one user to other users exhibiting different gaze behaviours on the same image stimuli.
By integrating the user embedding into a saliency prediction network~(\Cref{fig:teaser}), this additional input plays a crucial role in predicting filters that are convolved over extracted image features, thus integrating user-specific information essential for the personalised saliency prediction task. 
Our findings demonstrate the highly discriminative nature of these embeddings, enabling us to effectively compare individuals based on their distinctive gaze behaviour (\Cref{fig:TSNE_embeddings}).
Results on the downstream task of personal saliency prediction task show how the generated embeddings can be used to effectively
refine the universal saliency map predictions and tailor them to individual users. 
Moreover, we observe that these embeddings exhibit good generalisation capabilities when applied to both unseen users and images.~\footnote{Project code will be made publicly available upon acceptance.}

\section{Related Work}
\label{sec:related_work}

\subsection{Individual Differences in Visual Saliency}
Traditional saliency prediction methods ignore individual differences in visual salience between humans and instead predict an average, universal saliency map~\cite{borji2019salient}.
However, there are multiple prior works that show that humans have different visual preferences which draws their attention, which are stable and predictable.
De Haas and Linka et al.~\cite{de2019individual,linka2020osieshort} have identified multiple semantic dimensions along which human salience significantly differs, like faces.
Prior works have incorporated face detectors in saliency prediction pipelines, as they generally tend to attract significant attention~\cite{cerf2007predicting,borji2019salient}.
However, the results from De Haas et al. show that for specific humans, such predictions are imprecise as their attention is not attracted by faces.
Later Broda et al.~\cite{broda2022individual, broda2022individual2} identified that humans can be roughly clustered in two categories when observing persons. 
Either they tend to focus the head and inner facial features or they fixate on body parts like arms and legs.
Prior work has also studied which human traits influence the individual attention and found that for example age~\cite{krishna2018gaze} is an important factor as well as personality~\cite{hoppe2018eye} and gender~\cite{sammaknejad2017gender}.
Xu et al.~\cite{xu2018personalized} were the first that proposed a method to predict personalised instead of universal saliency by utilising such user traits as an additional input to their network.
However, it is still unclear which user traits are useful for saliency prediction, and explicitly collecting personal information might not be appropriate.
Moroto et al.~\cite{moroto2020few} later proposed a method involving a multi-task CNN to predict personalised saliency for each user in the training dataset. 
During inference, unseen users were matched to the seen users based on their similarity in attention allocation.
However, this requires many different users in the training data to generalise to unseen users.
Moreover, finding an appropriate similarity function to match unseen to seen users based on their gaze behaviour is challenging.
In contrast, we propose to train a single-task CNN that incorporates an additional user embedding as input, enabling the network to leverage user-specific information.

\subsection{User Embeddings}
User embeddings have found applications across diverse domains, serving either as a means to infer user-related information or to personalise the user experience.
In a seminal work by Pazzani~et~al.~\cite{pazzani1996syskill}, an approach was introduced where an agent learns a user embedding by leveraging explicit feedback provided by users in the form of page ratings.
This user embedding was then utilised to provide personalised website recommendations tailored to the user's interests and preferences.
Later, various methods have been proposed to construct user embeddings by incorporating diverse forms of application-specific explicit feedback~\cite{krulwich1997lifestyle, shavlik1998instructable}.
Methods that collect user information explicitly require the user to be active and might become inaccurate over time as the user's interest changes.

To address this limitation, implicit methods to create user embeddings have gained popularity. 
These methods allow users to simply interact with a system while the system creates a user embedding, without relying on explicit feedback~\cite{gauch2007user, eke2019survey}.
In a recent work by Wu et al.~\cite{wu2020author2vec}, a novel approach called Author2Vec was introduced to derive user embeddings by analysing the textual content authored by users on social media platforms. 
The authors demonstrated the superior performance of these embeddings in predicting user personality traits or mental states compared to alternative methods.
Similarly, An et al. \cite{an2021neural} used web browsing events to extract rich user embeddings for different downstream tasks.
In other related works, researchers investigated the incorporation of user embeddings as unique word tokens alongside the user's text.
This approach enables the model to comprehend the sentence in the context of the user embedding~\cite{zhong2021useradapter, mireshghallah2021useridentifier}.
In the eye-gaze domain He et al.~\cite{he2019device} have utilised appearance based user embeddings extracted from faces to personalise an appearance-based gaze estimator.
These works show the potential to leverage implicit user embeddings on a range of applications. 
In our work, to the best of our knowledge, we are the first to learn user embeddings from visual attentive behaviour and leverage them to enhance performance on the personalised saliency prediction task.

\begin{figure}[t]
    \centering
    \includegraphics[width=0.7\columnwidth]{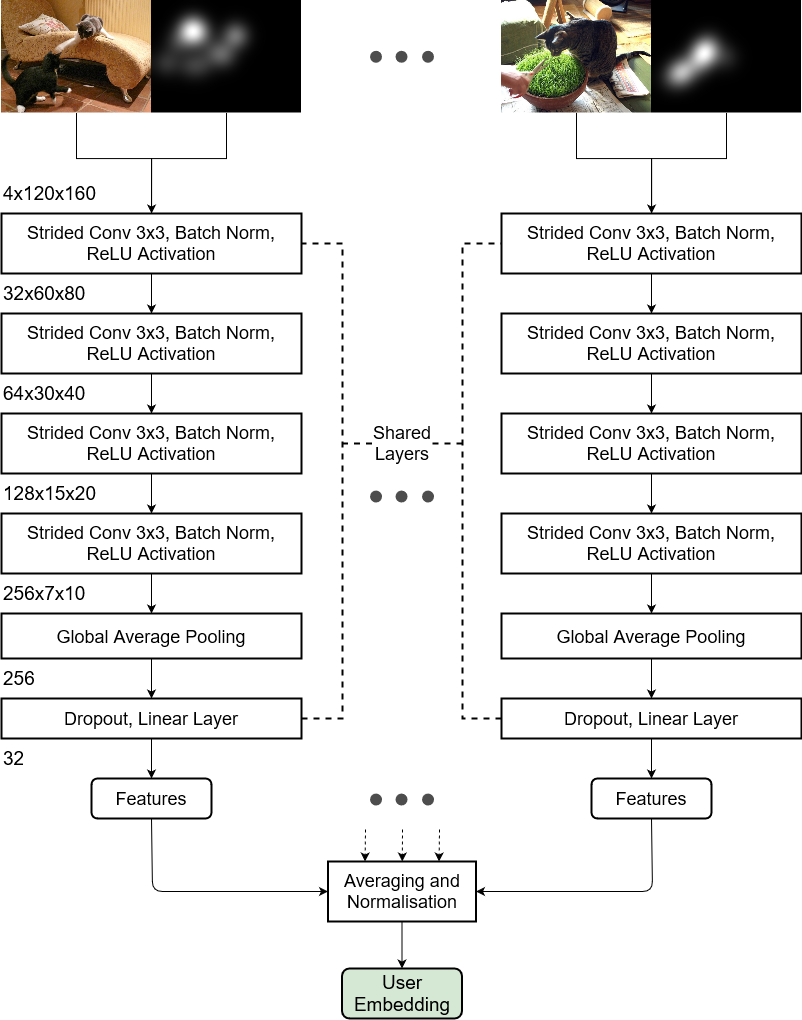}
    \caption{The architecture of our proposed user embedding extractor involves processing multiple images alongside an additional channel that includes the saliency information of a specific user, from which we aim to extract an embedding. To accomplish this, we employ a Siamese convolutional neural network, which is responsible for extracting features from each pair of image-saliency maps. Subsequently, the extracted features are averaged and normalised, resulting in the user embedding.}
    \label{fig:embedding_architecture}
\end{figure}

\section{Methodology}
\label{sec:method}
Traditional saliency prediction models learn a function $f(I) = \text{USM}_I$, where $I$ is an image and $\text{USM}_I$ is the corresponding \textit{Universal Saliency Map} (USM). The ground truth USM is calculated by averaging $n$ \textit{Personal Saliency Maps} (PSM) obtained from different humans observing the same image: $\text{USM}_I = \frac{1}{n} \sum_{k=1}^n \text{PSM}_{I,k}$.
Instead of predicting $\text{USM}_I$, our goal is to predict $\text{PSM}_{I,U}$ for a given image $I$ and user $U$.
To predict $\text{PSM}_{I,U}$ it is essential to incorporate additional input information that is specific to the user $U$ for whom the predictions are intended.

\subsection{Learning User Embeddings}
\label{sec:user_embeddings}

We propose a novel method for extracting user embeddings $e_U$ for each individual user $U$. 
These embeddings are derived from the distinct variations observed in the visual attention patterns of individual users.
Our hypothesis is that these embeddings can be effectively used by a personalised saliency prediction model to generate user-specific saliency outputs.

The architecture of our proposed embedding neural network is shown in \Cref{fig:embedding_architecture}.
A Siamese user embedding extractor $E$ takes $m$ different images $\{I_1,...,I_m\}$ along  with the corresponding PSMs $\{\text{PSM}_{I_1,U},...,\text{PSM}_{I_m,U}\}$ for the same user $U$ as input.
The goal is to extract joint image-saliency features which allow the network to understand the visual preferences of the user and extract a meaningful user embedding.
The PSM for each image is treated as an additional image channel besides the existing three RGB channels and resized to a resolution of $160\times 120$.
Each image-PSM tensor is passed through four convolution blocks consisting of a 2D convolution layer with stride two, a batch normalisation layer~\cite{ioffe2015batch} and a Rectified Linear Unit (ReLU) activation function.
Subsequently, a global average pooling layer~\cite{lin2013network} is employed to reduce the output to a one-dimensional vector, which helps prevent overfitting in conjunction with a dropout layer~\cite{srivastava2014dropout}. 
Finally, a linear layer predicts the output features for each image-PSM input.

This Siamese network is used to extract joint image-saliency features for each of the $m$ image-PSM pairs, which are subsequently averaged and normalised to unit length resulting in the extracted user embedding.
Preliminary experiments revealed that combining the extracted features from each image-PSM pair with a recurrent or transformer network results in severe overfitting.
Thus, averaging the features helps prevent overfitting and yields better results overall.

The network is optimised to minimise the triplet margin loss with online (semi-) hard triplet mining~\cite{hermans2017defense} defined as:
\begin{equation}
\label{eq:UE_loss}
\begin{aligned}
        &e_a = \left\|E(\{I_1,...,I_m\}, \{\text{PSM}_{I_1,U_1},...,\text{PSM}_{I_m,U_1}\})\right\|\\
        &e_p = \left\|E(\{I_{m+1},...,I_{2m}\}, \{\text{PSM}_{I_{m+1},U_1},...,\text{PSM}_{I_{2m},U_1}\})\right\|\\
        &e_n = \left\|E(\{I_{2m+1},...,I_{3m}\}, \{\text{PSM}_{I_{2m+1},U_2},...,\text{PSM}_{I_{3m},U_2}\})\right\|\\
        &\mathcal{L}_{E} = \textit{max}(e_a \cdot e_p - e_a \cdot e_n + m,0).
\end{aligned}
\end{equation}
To calculate the anchor user embedding $e_a$ in \Cref{eq:UE_loss}, a random user $U_1$ and $m$ random images $\{I_1,...,I_m\}$ with corresponding PSMs $\{\text{PSM}_{I_1,U_1},...,\text{PSM}_{I_m,U_1}\}$ are selected and passed through our embedding network. 
Similarly, the positive embedding example $e_p$ is calculated using the same user $U_1$ but with different images $\{I_{m+1},...,I_{2m}\}$ and PSMs $\{\text{PSM}_{I_{m+1},U_1},...,\text{PSM}_{I_{2m},U_1}\}$ , while the negative embedding example $e_n$ is obtained by selecting a different random user $U_2$, also with different images $\{I_{2m+1},...,I_{3m}\}$ and corresponding PSMs $\{\text{PSM}_{I_{2m+1},U_2},...,\text{PSM}_{I_{3m},U_2}\}$.
Based on the three embeddings $e_a$, $e_p$ and $e_n$ the standard triplet loss can be calculated as defined \Cref{eq:UE_loss}.
It is crucial to note that each image $I$ in the training dataset has a corresponding personalised saliency map ${\text{PSM}_{I,U}}$ for each user $U$. 
This ensures that the user embeddings are solely derived from the individual differences in users' visual attention behaviour, rather than being influenced by the images themselves. 
By minimising the distance between the anchor and the positive embedding, the network learns to recognise similar high-level gaze behaviour between different inputs for the same user. 
Similarly, maximising the distance between the anchor and the negative embedding encourages the network to distinguish the different attentive behaviour between two users.

While the training objective is to differentiate between different users, prior research has shown that embeddings learned through optimising the triplet loss have the capability to encode substantial class-specific information~\cite{hermans2017defense}. 
Additionally, similar classes tend to be close to each other in the embedding space, while dissimilar classes tend to have greater separation. 
This indicates that the embedding space effectively captures the relevant characteristics of the classes and presents a structured representation that reflects the underlying relationships between them.
By using the user embedding, a downstream task model can leverage the captured information and tailor its predictions to the specific characteristics and visual attention behaviour of each user.

\begin{figure}[t]
    \centering
    \includegraphics[width=0.9\columnwidth]{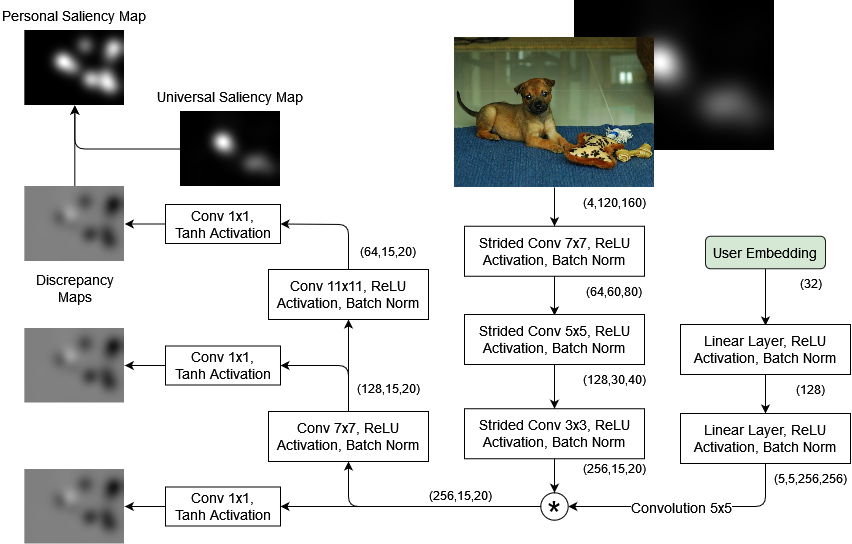}
    \caption{The personalised saliency map (PSM) network operates by taking an image stimulus and its corresponding universal saliency map (USM) as input. In addition, it incorporates user embedding, which is utilised to predict kernel weights, which are then convolved over the image-USM features. The network outputs a discrepancy map which can be added to the USM in order to generate the PSM.}
    \label{fig:PSM_architecture}
\end{figure}

\subsection{Personalised Saliency Prediction}

To demonstrate the effectiveness of the user embeddings, we adapted an existing method for personalised saliency prediction~\cite{xu2018personalized} and replace the manually, explicitly defined user characteristics with our learned embeddings from implicit gaze behaviour.

Similar to Xu et al~\cite{xu2018personalized} we define the PSM prediction task as a refinement of the USM:
\begin{equation}
\label{eq:PSM_def}
\text{PSM}(I,U) = \text{USM}(I) + \Delta(I,U),
\end{equation}
where $\text{PSM}(I,U)$ is the PSM of a user $U$ for a given image $I$, $\text{USM}(I)$ is the USM for that image and $\Delta(I,U)$ is the discrepancy map of that user and that image.
The discrepancy map essentially defines which parts of the image attract the specific users attention more or less compared to the average user.
Using this definition allows us to disentangle the prediction of the PSM, focusing on predicting the discrepancy map $\Delta(I,U)$ while using existing state-of-the-art models to predict the USM.
\Cref{fig:PSM_architecture} visualises our neural network architecture for personalised saliency prediction.
Input to the network $N$ are the image, the USM as a forth image channel and a user embedding.
First, a series of convolution layers extract image-saliency features.
In parallel, a series of linear layers predict filters based on the user embedding, which are subsequently convolved over the extracted image-saliency features.
This way the network can learn features that are specifically important to the user based on that users embedding~\cite{xu2018personalized}.
After a final sequence of convolution layers the model outputs a discrepancy map of size $15\times 20$ with a value range between -1 and 1.
We optimise the network with the mean squared error (MSE) loss between the predicted and the target discrepancy map.
Previous work has shown that by projecting extracted features from intermediate layers to the output space for additional supervision can improve performance \cite{cornia2016deep,lee2015deeply}.
Therefore we add an additional projection layer before the last two convolution layers to predict intermediate discrepancy maps.
Our overall loss objective $\mathcal{L}_{PSM}$ is then given as:
\begin{equation}
\label{eq:loss}
\mathcal{L}_{PSM} = \left\| \sum_{k=1}^3 N(\text{USM}(I), I, E_U)_k - \Delta(I,U) \right\|,
\end{equation}
where $N(...)_k$ is one of the three predicted discrepancy maps of the personalised saliency network.
To calculate the ground truth discrepancy maps $\Delta(I,U)$ we use \Cref{eq:PSM_def} and subtract the USM$_I$ from the PSM$_{I,U}$.

\section{Experiments}

\subsection{Implementation Details}
\paragraph{User Embedding Network.}
The architecture of the user embedding network is shown in \Cref{fig:embedding_architecture}.
The number of input image-PSM samples $m$ to our proposed Siamese CNN depends on the experiment and varies between 4 and 32.
Each model was optimised with the triplet margin loss as defined in ~\Cref{eq:UE_loss} with a margin of 0.05 and online semi-hard and hard triplet mining within a batch of 256 samples.
The weights were updated using the Adam optimiser~\cite{kingma2014adam} with a learning rate of 0.001 and default parameters otherwise.
The Dropout layer~\cite{srivastava2014dropout} masked neuron activations $50\%$ of the time during training.

\paragraph{Personalised Saliency Network.}
\Cref{fig:PSM_architecture} illustrates the architecture of our personalised saliency network. 
To effectively train this network, we first trained the embedding network to extract user embeddings, which serve as input. 
Since the embeddings rely on $m$ image-PSM pairs for their calculation, they can exhibit variability, particularly when $m$ is small. 
Hence, after training the embedding networks, we generated 100 embeddings for each participant within each dataset by randomly sampling data from the corresponding participant.
During the training process of the personalised saliency network, we randomly selected embeddings from this pool to account for these variations and improve generalisation. 
The models were trained with an SGD optimiser with initial learning rate of 0.02, momentum of 0.9, weight decay of 0.0005 and batch size 32.
The learning rate was reduced by a factor of 2 every 25 epochs. 
The trained model runs at 27 frames per second on an RTX 4070 GPU, achieving real-time personalised saliency predictions.

\paragraph{Evaluation Metrics}
We report multiple metrics commonly used to evaluate the similarity between saliency maps~\cite{pan2016shallow, xu2018personalized, judd2012benchmark}.
These metrics are the Pearson’s correlation coefficient (CC), similarity / histogram intersection (SIM), Area under ROC Curve (AUC-Judd)~\cite{judd2009learning}, normalised scanpath saliency (NSS) and Kullback-Leibler divergence (KLD).
In addition, to assess the accuracy of the user embeddings, we employ a labelling approach where an extracted embedding is considered correct if its nearest neighbour in the embedding space belongs to the same user. 
This evaluation metric is commonly referred to as precision at one~\cite{jarvelin2017ir}.

\begin{table}[t]
\fontsize{9}{10pt}\selectfont
\begin{center}
\begin{tabularx}{1.0\linewidth}{l *5{>{\centering\arraybackslash}X}}
\toprule
Model & CC & SIM & AUC & NSS & KLD \\
\cmidrule(lr){1-1} \cmidrule(lr){2-2} \cmidrule(lr){3-3} \cmidrule(lr){4-4} \cmidrule(lr){5-5} \cmidrule(lr){6-6}
DeepGazeIIE (DG) & 0.622 & 0.556 & 0.904 & 2.121 & 1.123 \\
Fine-tuned DG & 0.715 & 0.619 & \underline{0.905} & 2.140 & \underline{0.526} \\
MultiCNN w/ DG & \underline{0.735} & \underline{0.643} & 0.897 & \underline{2.142} & 0.708 \\
\cmidrule(l){2-6}
Ours w/ DG & \textbf{0.736} & \textbf{0.651}$^*$ & \textbf{0.907} & \textbf{2.170}$^*$ & \textbf{0.509}$^*$ \\
\midrule
Ground Truth USM & 0.801 & \underline{0.685} & \underline{0.921} & 2.373 & \underline{0.372} \\
MultiCNN w/ GT & \underline{0.804} & 0.683 & 0.919 & \underline{2.378} & 0.511 \\
\cmidrule(l){2-6}
Ours w/ GT & \textbf{0.813}$^*$ & \textbf{0.706}$^*$ & \textbf{0.922} & \textbf{2.405}$^*$ & \textbf{0.357}$^*$ \\
\bottomrule
\end{tabularx}
\end{center}
\caption{Closed-set results for the ID dataset~\cite{de2019individual}. * indicates significant improvement over the strongest baseline.}
\label{tbl:results_ID_CS}
\end{table}

\begin{table}[t]
\fontsize{9}{10pt}\selectfont
\begin{center}
\begin{tabularx}{1.0\linewidth}{l *5{>{\centering\arraybackslash}X}}
\toprule
Model & CC & SIM & AUC & NSS & KLD \\
\cmidrule(lr){1-1} \cmidrule(lr){2-2} \cmidrule(lr){3-3} \cmidrule(lr){4-4} \cmidrule(lr){5-5} \cmidrule(lr){6-6}
DeepGazeIIE (DG) & 0.584 & 0.530 & 0.881 & 2.288 & 1.294 \\
Fine-tuned DG & 0.734 & 0.622 & 0.887 & 2.293 & \underline{0.562} \\
MultiCNN w/ DG & \underline{0.746} & \underline{0.637} & \underline{0.892} & \underline{2.299} & 0.604 \\
\cmidrule(l){2-6}
Ours w/ DG & \textbf{0.760}$^*$ & \textbf{0.649}$^*$ & \textbf{0.896} & \textbf{2.308} & \textbf{0.481}$^*$ \\
\midrule
Ground Truth USM & 0.821 & 0.698 & 0.912 & 2.500 & \underline{0.366} \\
MultiCNN w/ GT & \underline{0.845} & \underline{0.711} & \underline{0.914} & \textbf{2.609} & 0.415 \\
\cmidrule(l){2-6}
Ours w/ GT & \textbf{0.846} & \textbf{0.725}$^*$ & \textbf{0.915} & \underline{2.595} & \textbf{0.336}$^*$ \\
\bottomrule
\end{tabularx}
\end{center}
\caption{Closed-set results for the PS dataset~\cite{xu2018personalized}. * indicates significant improvement over strongest baseline.}
\label{tbl:results_PS_CS}
\end{table}

\subsection{Datasets}
\label{sec:datasets}

Saliency prediction models are typically pre-trained using the large-scale SALICON dataset~\cite{jiang2015salicon}.
However, we cannot utilise the SALICON dataset for learning the embeddings as 
each subject from the dataset observed a different subset of image stimuli.
As discussed in \Cref{sec:user_embeddings}, it is critical that each participant looks at all images, or at least that there is a significant overlap between participants, as otherwise the model can simply identify the user based on the images they observed ignoring their specific visual attention behaviour.

Based on the above requirement, we selected two publicly available datasets where each participant observed each image while their gaze was recorded with an eye-tracker.
The first dataset was collected by Xu et al.~\cite{xu2018personalized}, which we call the \textit{Personalised Saliency} (PS) dataset.
The PS dataset contains 1,600 images, which they selected to contain many different semantic categories in each image, as they argue that this maximises the variation of visual attention between participants.
Each stimuli in the dataset was observed by 30 participants for three seconds a total of four times, allowing them to average out stochastic variations in each participants attentive behaviour.
To evaluate how well our system generalises to unseen images we split the images into $80\%$ for training, $10\%$ for validation and $10\%$ for testing.
Furthermore, to evaluate how well our system generalises to unseen participants, we split the 30 participants into 20 for training and 5 each for validation and testing.

The second dataset was collected by Haas et al.~\cite{de2019individual} which we call the \textit{Individual Differences} (ID) dataset.
The stimuli were selected to be comprised of complex scenes containing multiple different semantic categories and objects.
They recorded gaze data from a total of 102 different participants each looking at 700 different image stimuli.
Similar to the PS dataset we split the images into $80\%$ for training, $10\%$ for validation and $10\%$ for testing.
Furthermore, we split the 102 participants into 80 for training, 10 for validation and 12 testing.

Following Xu et al.~\cite{xu2018personalized} we conduct both closed-set and open-set experiments for each dataset. 
In the closed-set experiments, the same participants were used in both the training, validation, and test sets with the validation and test sets containing unseen images.
This enables us to assess the capability to predict personalised saliency for familiar users. 
As for the open-set experiments, we assess the personalised saliency prediction performance on unseen participants from the test set. 
This analysis helps us understand how effectively models can generalise to new users.

\subsection{Baselines}
We evaluate our method against three baselines. 
Firstly, we utilise DeepGaze IIE~\cite{linardos2021deepgaze}, a state-of-the-art universal saliency prediction model. 
We compare the predicted USMs by DeepGaze IIE with our refined PSMs based on the DeepGaze IIE prediction. 
Since DeepGaze IIE was not originally trained on our dataset, we fine-tune the model's prediction on each dataset and present results both with and without fine-tuning.
Secondly, we compare the ground truth USMs with our predicted PSMs generated through refining the USMs.

In the literature, there are two relevant works by Xu et al.~\cite{xu2018personalized} and Moroto et al.~\cite{moroto2020few} that propose methods for PSM prediction, as discussed in Section \ref{sec:related_work}. 
Unfortunately, we were unable to directly compare our results with Xu et al. due to the unavailability of the user-specific information required for their method. 
However, for the closed set experiments they also proposed MultiCNN, where they train a separate classifier for each participant, which we will use as our third baseline.
The MultiCNN architecture is identical to our network shown in \Cref{fig:PSM_architecture} without the embedding pathway.

Moroto et al.~\cite{moroto2020few} propose a method to map unseen users to the training user with the most similar visual attention behaviour, which allows them to make personalised saliency predictions with the corresponding trained classifier.
Inspired by this, we propose an oracle mapping by evaluating unseen participants using every trained MultiCNN model and then choose the model achieving the lowest loss.
This allows us to provide the upper-bound MultiCNN performance for the open set experiments.

\begin{table}[t]
\fontsize{9}{10pt}\selectfont
\begin{center}
\begin{tabularx}{1.0\linewidth}{l *5{>{\centering\arraybackslash}X}}
\toprule
Model & CC & SIM & AUC & NSS & KLD \\
\cmidrule(lr){1-1} \cmidrule(lr){2-2} \cmidrule(lr){3-3} \cmidrule(lr){4-4} \cmidrule(lr){5-5} \cmidrule(lr){6-6}
DeepGazeIIE (DG) & 0.615 & 0.553 & \underline{0.916}  & 2.239 & 1.083 \\ 
Fine-tuned DG & 0.721 & 0.624 & 0.914 & 2.244 & \underline{0.542} \\ 
MultiCNN w/ DG & \underline{0.723} & \underline{0.627} & 0.910 & \underline{2.244} & 0.620 \\
\cmidrule(lr){2-6}
Ours w/ DG & \textbf{0.726} & \textbf{0.638}$^*$ & \textbf{0.916} & \textbf{2.253} & \textbf{0.527}$^*$ \\ 
\midrule
Ground Truth USM & 0.804 & 0.682 & 0.922 & 2.506 & \underline{0.387} \\ 
MultiCNN w/ GT & \underline{0.804} & \underline{0.690} & \underline{0.929} & \underline{2.550} & 0.440 \\
\cmidrule(lr){2-6}
Ours w/ GT & \textbf{0.813}$^*$ & \textbf{0.702}$^*$ & \textbf{0.933} & \textbf{2.552} & \textbf{0.366}$^*$ \\ 
\bottomrule
\end{tabularx}
\end{center}
\caption{Open-set results for the ID dataset~\cite{de2019individual}. * indicates significant improvement over strongest baseline.}
\label{tbl:results_ID_OS}
\end{table}

\begin{table}[t]
\fontsize{9}{10pt}\selectfont
\begin{center}
\begin{tabularx}{1.0\linewidth}{l *6{>{\centering\arraybackslash}X}}
\toprule
Model & CC & SIM & AUC & NSS & KLD \\
\cmidrule(lr){1-1} \cmidrule(lr){2-2} \cmidrule(lr){3-3} \cmidrule(lr){4-4} \cmidrule(lr){5-5} \cmidrule(lr){6-6}
DeepGazeIIE (DG) & 0.564 & 0.522 & 0.860 & 1.944 & 1.431 \\ 
Fine-tuned DG & 0.692 & \underline{0.618} & 0.862 & \underline{2.032} & 0.685 \\ 
MultiCNN w/ DG & \underline{0.693} & 0.615 & \underline{0.862} & 2.006 & \underline{0.682} \\
\cmidrule(lr){2-6}
Ours w/ DG & 0.712 & 0.634 & 0.870 & 2.033 & 0.539 \\ 
Ours CD w/ DG & \textbf{0.713}$^*$ & \textbf{0.634}$^*$ & \textbf{0.871}$^*$ & \textbf{2.033} & \textbf{0.533}$^*$ \\ 
\midrule
Ground Truth USM & 0.788 & 0.691 & 0.892 & 2.218 & \underline{0.399} \\ 
MultiCNN w/ GT & \underline{0.800} & \underline{0.694} & \underline{0.892} & \textbf{2.290} & 0.516 \\
\cmidrule(lr){2-6}
Ours w/ GT & 0.804 & 0.701 & 0.893 & 2.273 & 0.371 \\ 
Ours CD w/ GT & \textbf{0.807} & \textbf{0.704}$^*$ & \textbf{0.894} & \underline{2.288} & \textbf{0.361}$^*$ \\ 
\bottomrule
\end{tabularx}
\end{center}
\caption{Open-set results for the PS dataset~\cite{xu2018personalized}. * indicates significant improvement over strongest baseline.}
\label{tbl:results_PS_OS}
\end{table}

\begin{figure*}[t]
    \centering
    \includegraphics[width=\linewidth]{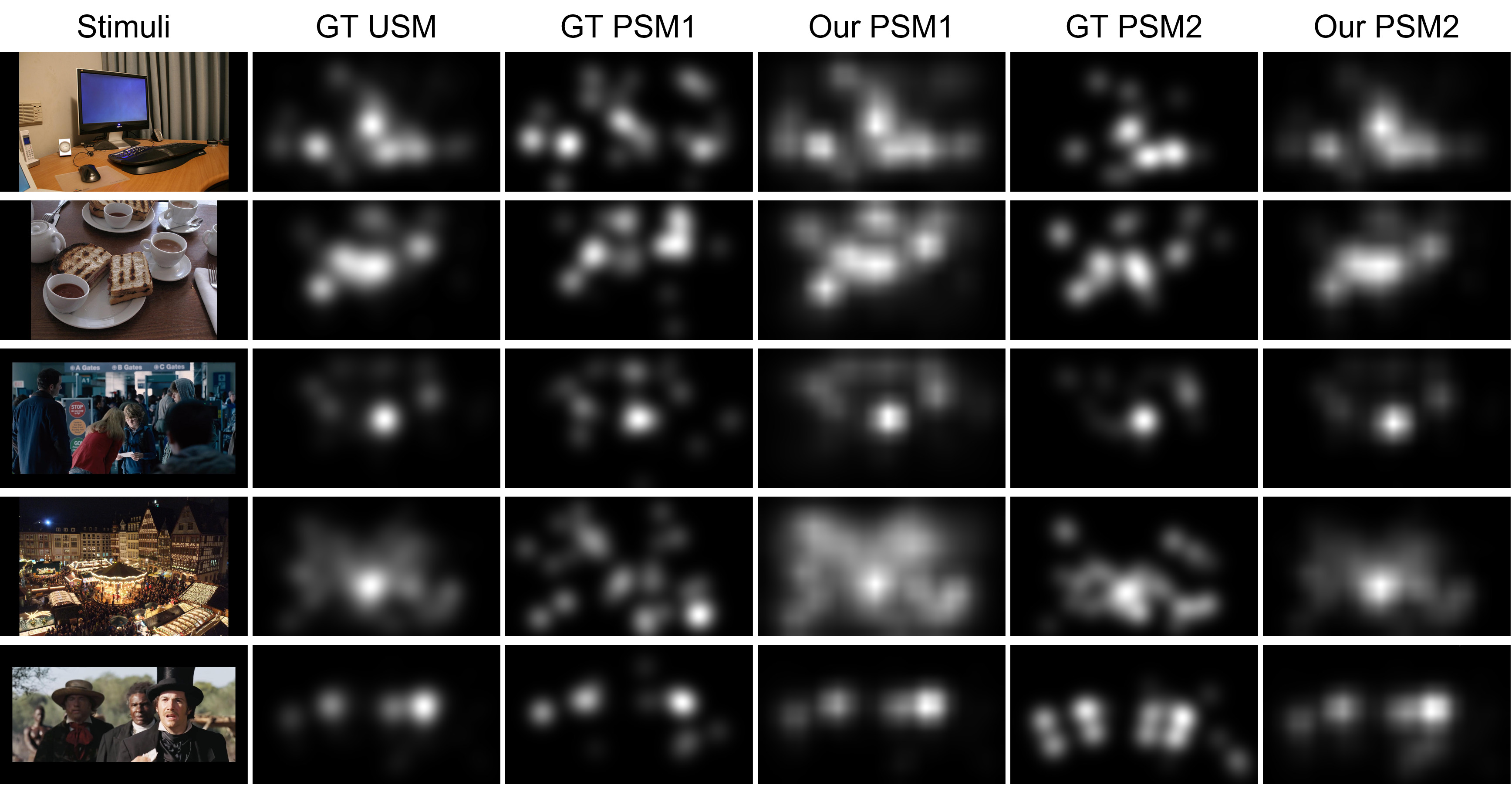}
    \caption{Example PSM predictions for two users from the PS~\cite{xu2018personalized} test set with our proposed method compared to the ground truths.}
    \label{fig:example_PSM_predictions}
\end{figure*}

\subsection{Personalised Saliency Prediction}
\paragraph{Closed-Set Results}
\Cref{tbl:results_ID_CS} shows the closed-set results for the ID dataset and \Cref{tbl:results_PS_CS} shows the results for the PS dataset.
The results with our method were obtained using embeddings extracted from $m=32$ image-PSM pairs for each user.
The best performing models are highlighted in bold, while the performance of the second best model is underlined.
Significance tests were conducted using the Mann-Whitney-U test for each metric with a p-value threshold of $<0.05$.
An asterisk next to metrics for our method indicates that the improvement compared to the strongest baseline was statistically significant.
The ground truth USM is the upper bound traditional saliency prediction methods could potentially achieve.
The results for MultiCNN and Ours show that it is possible to further refine these USMs as overall both methods outperform the ground truth USM baseline.
Furthermore, we observe that for both datasets our method outperforms all baselines including MultiCNNs in all metrics except for NSS on the PS dataset.
However, for a real-world scenario the ground truth USM might not be available and has to be predicted first.
The results show that MultiCNNs and Ours using the non fine-tuned USM predictions from DeepGaze IIE still outperform the fine-tuned DeepGaze IIE baseline, with our method achieving the best performance.
This indicates that our method can be applied to different potential suboptimal USM predictions and still produce a more refined PSM prediction.

\paragraph{Open-Set Results}

\Cref{tbl:results_ID_OS} shows the open-set results for the ID dataset and \Cref{tbl:results_PS_OS} shows the results for the PS dataset.
We can observe a very similar trend as with the closed-set experiments with our method overall outperforming all baselines, indicating that our embeddings help the personal saliency network to generalise well to unseen participants.
Since the PS dataset consists only of 30 different participants of which 20 are used for training, learning generalisable user embeddings is more challenging.
We therefore experiment with combining the training splits of the PS and ID datasets when training the user embedding network, allowing the network to observe the attentive behaviour of a total of 100 different participants during training.
Note that we still trained and evaluated our personalised saliency network using only the PS dataset. 
We report the resulting performance on the combined dataset (CD) as \textit{Ours CD} in \Cref{tbl:results_PS_OS}.
We can observe that using these new embeddings the performance of the personalised saliency model further improves for all metrics.

In addition to the quantitative results, \Cref{fig:example_PSM_predictions} shows example open-set PSM predictions for multiple image stimuli on the test split of the PS dataset.
In the provided examples, it is evident that the user represented in the last two columns exhibits a more focused attention behaviour compared to the first user. 
Our embeddings successfully capture this distinction, as indicated by the corresponding saliency predictions.
Notably, in the example depicted in the bottom row, the second user appears to allocate less attention to faces and instead also focuses on other body parts. 
Our embeddings seem to capture this behaviour, as evidenced by the saliency allocation below the face region in the corresponding PSM predictions.
Together with our quantitative results this further demonstrates the effectiveness of using user embeddings learned from visual attention for more personalised saliency predictions.

\begin{figure}[t]
    \centering
    \includegraphics[width=0.6\columnwidth]{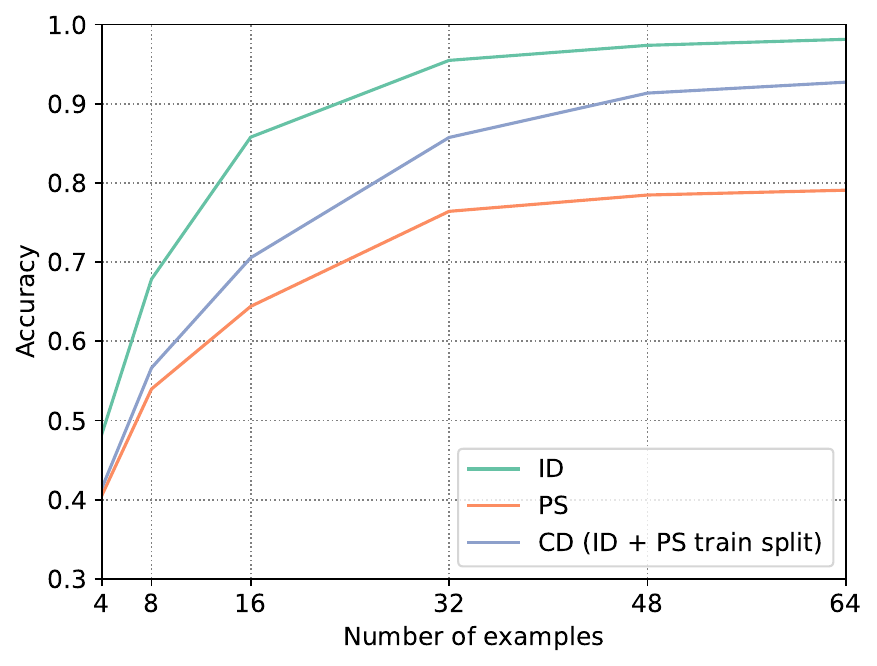}
    \caption{Performance comparison of different user embedding extraction models. The y-axis indicates the model's accuracy and the x-axis how many image-PSM examples $m$ were used as input to extract the embedding (4, 8, 16, 32, 48 or 64). We report the accuracy for unseen participants on the Individual Differences (ID) dataset, the Personal Saliency (PS) dataset and for the combined CD dataset.}
    \label{fig:embedding_accuracy}
\end{figure}

\subsection{User Embeddings Analysis}
To gain a better understanding of our extracted user embeddings, we further analyse the performance of the user embedding network.
\paragraph{Number of Examples for Embedding Extraction}
\Cref{fig:embedding_accuracy} shows the model's test set accuracy on the y-axis for different datasets and the number of examples $m$ used for embedding extraction on the x-axis.
We report the test set accuracy for both datasets ID and PS, as well as the PS test set accuracy when training on the combined training set CD.
The user embedding network achieves an accuracy of $98.1\%$ on the ID test split when using $m=64$ examples, showing that it is able to differentiate very well between participants that the model never saw during training.
Similarly the model trained on the PS dataset achieves an accuracy of $79.1\%$ when only using the PS training split and an accuracy of $92.7\%$ when combining the training splits from PS and ID.
As already indicated by the improved personalised saliency prediction results reported in \Cref{tbl:results_PS_OS}, training on both datasets increases the performance on the PS dataset, which only contains a small number of participants.
Note that the random baseline accuracy for the ID dataset is $8.3\%$ (12 participants in the test set), while it is $20\%$ for the PS dataset (5 participants in the test set).
Furthermore, we can observe that the model's accuracy increases with the number of examples $m$ provided as input.
Combining the training datasets especially improves the accuracy for larger $m$.
This is likely due to the increased user information the model can extract with larger $m$, resulting in better generalisation if provided with more diverse users during training.

\begin{figure}[t]
\centering
\begin{subfigure}[b]{0.5\linewidth}
  \centering
  \includegraphics[width=0.6\linewidth]{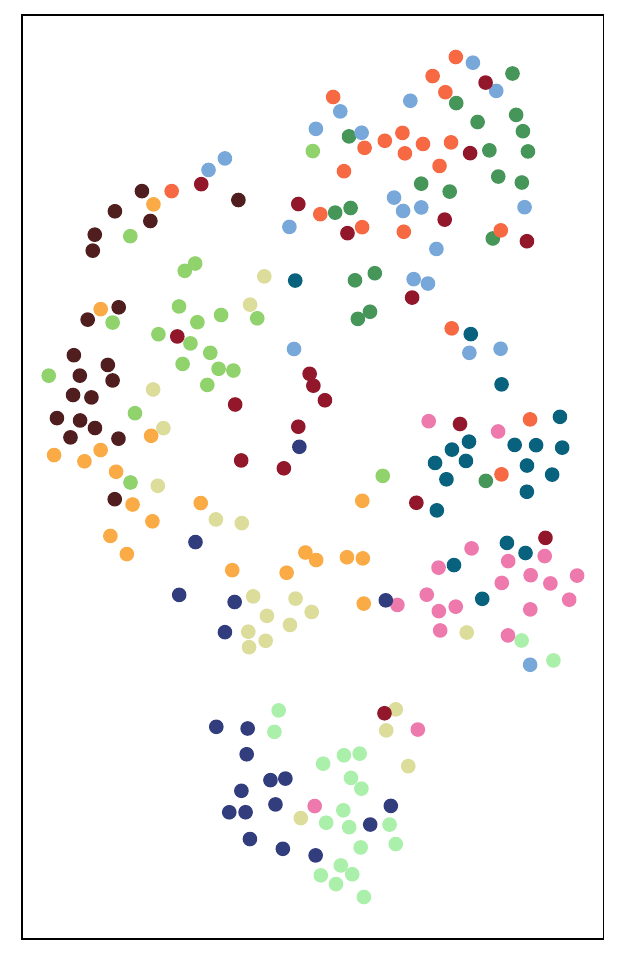}
  \caption{}
  \label{fig:ours_thumbnail}
\end{subfigure}\hfill
\begin{subfigure}[b]{0.5\linewidth}
  \centering
  \includegraphics[width=0.6\linewidth]{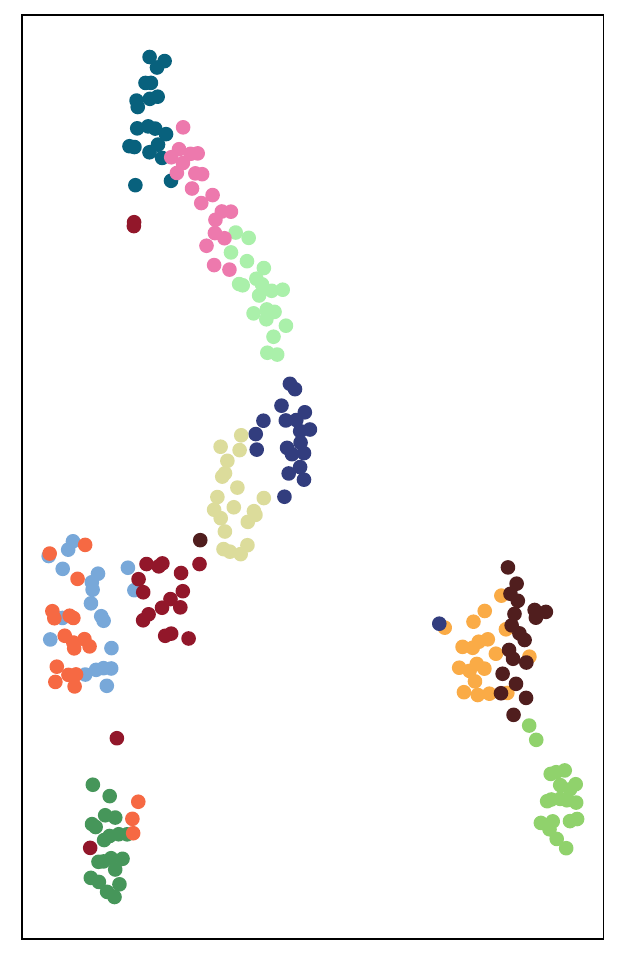}
  \caption{}
  \label{fig:GAN_thumbnail}
\end{subfigure}
\caption{We use t-SNE to reduce the 32-dimensional embeddings into two dimensions. Figure (a) shows the embedding space for embeddings extracted using $m=8$ image-PSM pairs while in (b) we used $m=32$ pairs. Each dot represents one embedding extracted using $m$ randomly sampled image-PSM pairs. Each colour corresponds to a unique user from the ID test set.}
\label{fig:TSNE_embeddings}
\end{figure}

\paragraph{Visualisation of the Embedding Space}
To analyse the embedding space we reduced the 32 dimensional user embeddings to 2 dimensions using t-SNE.
\Cref{fig:TSNE_embeddings} illustrates the 2-dimensional embeddings for the 12 participants from the ID test split. 
In this visualisation, each dot represents an embedding, and the colour of each dot corresponds to a specific user.
We calculated 20 embeddings for each participant by randomly sampling $m$ input image-PSM pairs.
In (a) we visualise the embeddings for $m = 8$ while (b) shows the embeddings for $m = 32$.
We observe that the embeddings for $m=8$ have a much larger variance within a participant compared to the embeddings extracted with $m=32$, as less information about the user can be extracted.
For a low $m$ the boundaries between participants becomes fuzzy which is also reflected by the lower accuracy reported in \Cref{fig:embedding_accuracy}.
We can observe that for $m=32$ the embeddings are highly discriminative as clear user clusters are formed.

\section{Broader Impact}
Our method demonstrates the ability to extract user embeddings from a small amount of gaze data, showcasing that these embeddings effectively capture relevant information for modelling personal saliency.
Predicting saliency is a crucial task with implications for various downstream applications in computer vision and human-computer interaction. 
For instance, attentive user interfaces aim to manage the user's attention effectively, relying on knowledge about their visual attention~\cite{vertegaal2002designing}. 
Recommender systems can benefit from understanding users' visual attention to enhance the visibility of top-ranked entities~\cite{zhao2016gaze}. 
Other downstream tasks that benefit from accurate personalised saliency include video summarization~\cite{ma2002user}, automated image cropping~\cite{ardizzone2013saliency}, and image captioning~\cite{he2019human}.
While the downstream task of this work is personalised saliency prediction, our proposed method for extracting user embeddings could prove advantageous for other downstream tasks where personalisation is crucial.

Although personalised saliency is helpful for many tasks, the underlying computational user model could also be misused to synthesise data for a given user by impersonating the user's own saliency.
Furthermore, instead of using the embeddings to predict personal saliency, they might be misused to directly extract user sensitive information.
These embeddings might encode user characteristics or private information about the user that correlates with their gaze behaviour. 
For example, prior work has shown that gender is a strong modulator of saliency preference~\cite{hewig2008gender}, and as such, this private information might be extractable from our embeddings.
This encoded information could be used to identify users based on their personal saliency, especially as attention tracking becomes more pervasive and cheap through appearance based gaze estimation or mouse tracking.
Another factor to consider is that our embeddings may be biased as they are extracted from a small number of stimuli with corresponding gaze data.
As we can see in \Cref{fig:TSNE_embeddings}, the embeddings vary for the same person and we can notice multiple outlier embeddings that are far apart form their cluster, even when using more data.
Thus, when used for tasks like user profiling this might lead to incorrect user profiles and subsequently incorrect predictions in downstream tasks.

Continuing this line of research, it is conceivable that future work will not only synthesise personalised saliency but even raw gaze of specific users with such embeddings.
Accurate computational models of user-specific gaze behaviour would be of great significance for even more downstream tasks. 
However, this advancement might raise potential security risks for applications that fundamentally rely on gaze behaviour analysis, such as gaze-based authentication.

\section{Summary}
In this work, we proposed a novel method that extracts user embeddings from pairs of natural images and corresponding user-specific saliency maps.
The learned embeddings capture the users' unique characteristics and can be used to address the personalised saliency prediction task. 
In contrast to prior work for this task that required explicit user input, our method only requires implicit input from gaze behaviour collected using an eye tracker. 
Our proposed method uses a Siamese convolutional neural encoder to learn the embedding model, trained by contrasting a user's gaze behaviour with that of different users.
Results on two saliency datasets demonstrated the embeddings' discriminative power, our method's generalisability to unseen users and images, and improved performance over universal saliency prediction models. 
As such, our work presents a promising approach to learning and leveraging user embeddings from implicit behaviour also for other tasks or applications that require individual user characteristics.

\begin{acks}
Florian Strohm and Andreas Bulling were funded by the European Research Council (ERC) under the grant agreement 801708.
Mihai Bâce was funded by a Swiss National Science Foundation (SNSF) Postdoc.Mobility Fellowship (grant number 214434).
\end{acks}

\bibliographystyle{ACM-Reference-Format}
\bibliography{main}

\end{document}